\documentclass[letterpaper, 10 pt, conference]{ieeeconf}  

\IEEEoverridecommandlockouts                              

\usepackage{graphics} 
\usepackage{epsfig} 
\usepackage{amsmath} 
\usepackage{amssymb}  

\usepackage{fancyhdr}
\usepackage{xcolor}

\title{\LARGE \bf
LLM-based ambiguity detection in natural language instructions for collaborative surgical robots}

\author{Ana Davila$^{1}$ and Jacinto Colan$^{2}$ and Yasuhisa Hasegawa$^{1}$
\thanks{$^{1}$\quad
Institutes of Innovation for Future Society, Nagoya University, Furo-cho, Chikusa-ku, Nagoya, Aichi 464-8601, Japan}
\thanks{Correspondence: {\tt\small davila.ana@robo.mein.nagoya-u.ac.jp}}
\thanks{$^{2}$\quad
Department of Micro-Nano Mechanical Science and Engineering, Nagoya University, Furo-cho, Chikusa-ku, Nagoya, Aichi 464-8603, Japan}
\thanks{
This work was supported in part by the Japan Science and Technology Agency (JST) CREST including AIP Challenge Program under Grant JPMJCR20D5, and in part by the Japan Society for the Promotion of Science (JSPS) Grants-in-Aid for Scientific Research (KAKENHI) under Grant 25K21247.}
}

\fancypagestyle{firstpage}{
  \fancyhf{}
  
  \fancyhead[L]{
    \vspace*{-25pt} 
    \parbox{\textwidth}{
      \textcolor{blue}{\small This work has been accepted at the 2025 IEEE International Conference on Robot and Human Interactive Communication (ROMAN) and submitted to the IEEE for possible publication. Copyright may be transferred without notice, after which this version may no longer be accessible.}
    }
  }
}

\begin{document}

\maketitle
\thispagestyle{firstpage}
\pagestyle{empty}

\begin{abstract}
Ambiguity in natural language instructions poses significant risks in safety-critical human-robot interaction, particularly in domains such as surgery. To address this, we propose a framework that uses Large Language Models (LLMs) for ambiguity detection specifically designed for collaborative surgical scenarios. Our method employs an ensemble of LLM evaluators, each configured with distinct prompting techniques to identify linguistic, contextual, procedural, and critical ambiguities. A chain-of-thought evaluator is included to systematically analyze instruction structure for potential issues. Individual evaluator assessments are synthesized through conformal prediction, which yields non-conformity scores based on comparison to a labeled calibration dataset. Evaluating Llama 3.2 11B and Gemma 3 12B, we observed classification accuracy exceeding 60\% in differentiating ambiguous from unambiguous surgical instructions. Our approach improves the safety and reliability of human-robot collaboration in surgery by offering a mechanism to identify potentially ambiguous instructions before robot action.
\end{abstract}

\section{Introduction}

The integration of robotic systems into high-stakes environments, particularly surgical settings, has accelerated the need for reliable human-robot communication. Natural language instructions offer an intuitive interface between healthcare professionals and robotic assistants, but the inherent ambiguity of natural language presents significant challenges. In surgical contexts, where precision is essential, instruction ambiguity can lead to potentially dangerous misinterpretations, compromising patient safety and surgical outcomes \cite{satchidanand21put}. These ambiguities manifest in multiple forms, including linguistic, contextual, and procedural ambiguities, each creating an opportunity for critical misunderstanding.

Traditional approaches to ambiguity management in human-robot instruction often rely on rule-based systems, predefined ontologies, or simple clarification dialogues \cite{villamar21ontology}. While functional in controlled environments with limited instruction variability, these methods frequently struggle within the dynamic and complex nature of surgical scenarios. Surgical instructions often contain domain-specific terminology, rely on implicit knowledge, and carry high stakes for misinterpretation, exceeding the capabilities of many conventional systems. Furthermore, these approaches may fail to capture the full spectrum of ambiguity types encountered in realistic operational settings.

Recent advancements in Large Language Models (LLMs) demonstrate remarkable capabilities in natural language understanding, context sensitivity, and nuanced reasoning. Their ability to process complex language and identify subtle semantic relationships suggests significant potential for overcoming the limitations of rule-based systems in ambiguity detection. While LLMs have been explored for enhancing robot task planning \cite{singh23progprompt} and improving human-robot collaborative workflows \cite{zhang23large}, their specific application to systematic ambiguity detection in safety-critical surgical instructions remains relatively underexplored.

\begin{figure}[t]
  \centering
  \includegraphics[width=0.9\linewidth]{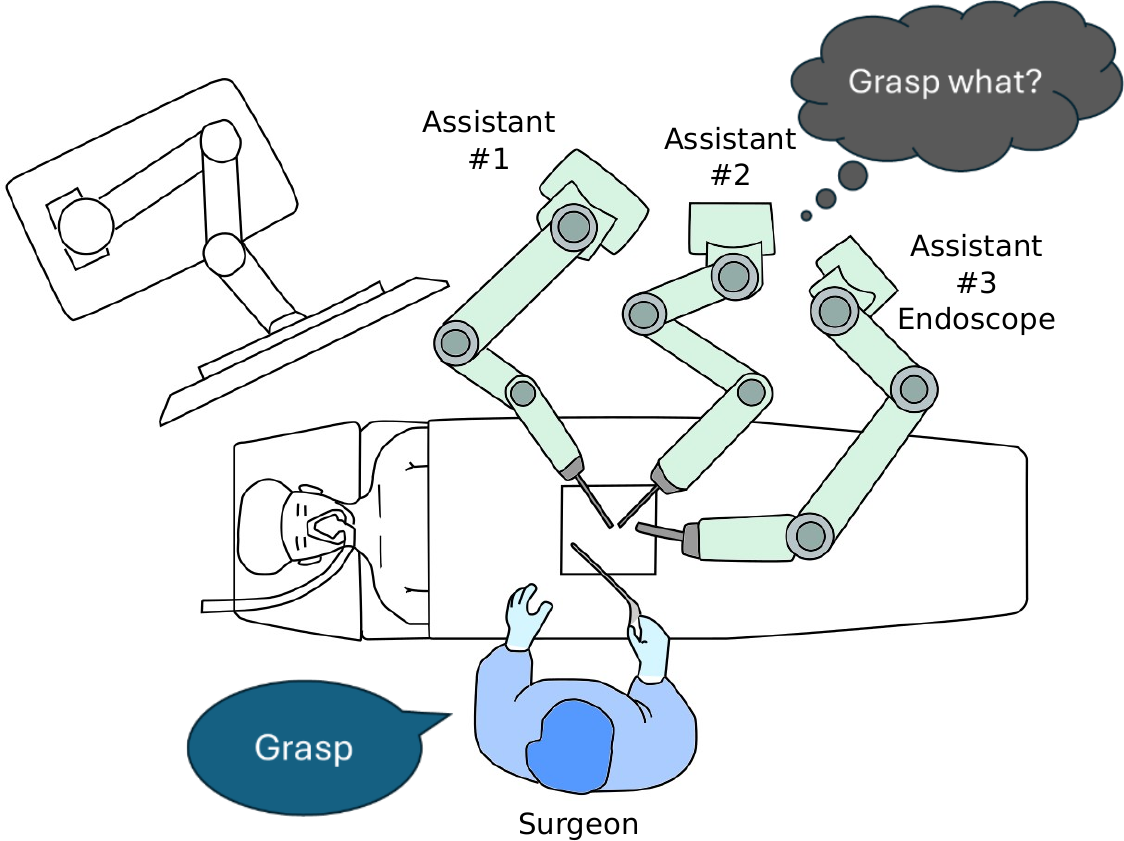}
  \caption{Collaborative robot-assisted surgery requires seamless communication between surgeons and robotic assistants.}
  \label{fig:1}
\end{figure}

This paper introduces a novel framework for detecting ambiguity in natural language instructions for surgical robotic assistants as shown in Figure~\ref{fig:1}. Our approach leverages an ensemble of specialized LLM-based evaluators, each designed to identify distinct ambiguity types through targeted prompting strategies. By combining these diverse perspectives, we aim to comprehensively capture multiple dimensions of potential instruction ambiguity. To transform these evaluations into reliable ambiguity classifications, we implement a conformal prediction methodology that provides statistical guarantees about classification confidence. This approach allows a robotic system to quantify uncertainty in instruction interpretation and appropriately determine when clarification is necessary before execution.

The main contributions of this paper are:

\begin{itemize}
\item Development of an ensemble of LLM evaluators for surgical robotics to address various ambiguities.
\item Use of conformal prediction to quantify ambiguity via calibrated scores.
\item Demonstration of the approach's effectiveness with modern LLMs in recognizing ambiguous instructions.
\end{itemize}

\section{Related Works}

\subsection{Ambiguity in Natural Language}

Natural language ambiguity presents significant challenges for human-robot interaction systems, particularly in contexts requiring precise instruction interpretation. The problem of ambiguity has been well studied in computational linguistics and natural language processing (NLP). Foundational work in NLP categorizes ambiguities (e.g., lexical, syntactic, semantic, pragmatic) \cite{gleich10ambiguity} and highlights challenges such as referential ambiguity, where context is needed to resolve which object an instruction refers to \cite{yang11analysing}. Recent work explores the use of LLMs themselves to handle ambiguity, for example, aligning LLM outputs with ambiguous inputs through self-assessment of uncertainty \cite{kim24aligning} or using structured taxonomies to guide instruction refinement \cite{niwa24ambignlg}. These studies provide core concepts and LLM-based techniques relevant to the interpretation of potentially ambiguous robotic instructions.

\subsection{Ambiguity Detection in Robotic Systems}

The robotics community has developed specialized approaches for handling unclear instructions across various domains. Wang et al. \cite{wang25learning} present an Ask-when-Needed framework where LLMs evaluate instruction clarity before execution, seeking clarification only when necessary. Other studies integrate LLM with perception (vision-language models) to resolve object reference ambiguities during tasks such as manufacturing or collaborative assembly \cite{fan24vision}. Interactive systems like SeeAsk strategically query users when visual grounding is uncertain during grasping tasks \cite{mo23towards}. These approaches demonstrate the value of LLMs in identifying and sometimes resolving ambiguity in general robotic contexts, often through interaction or multi-modal sensing.

However, in the surgical context, the risks associated with ambiguity are amplified due to its high stakes, specialized terminology, dynamic environment, and need for extreme precision. Studies have highlighted the difficulty in translating qualitative voice commands (for example, "move more left") into precise robotic actions, especially with changing visual fields during procedures such as endoscopy \cite{zinchenko17study}. Although several natural language interfaces for surgical robots have been developed \cite{elazzazi22natural, davila24voice, moghani24sufia}, improving usability, they often lack dedicated mechanisms for systematic detection of ambiguities. Some research implicitly addresses ambiguity through multi-modal inputs like gesture control \cite{jacob12gestonurse} or utilizes multi-modal LLMs for specific surgical sub-tasks like interpreting context for robotic assistance in blood suction \cite{zargarzadeh25fromdecision}. However, a comprehensive framework specifically targeting the detection of diverse ambiguity types within natural language instructions for surgical robots remains an open area.
 
To move beyond simple detection towards reliable decision-making (i.e., deciding whether to execute or clarify), methods for quantifying uncertainty are essential. Conformal Prediction (CP) \cite{angelopoulos22gentle} has emerged as a promising statistical framework for providing rigorous confidence bounds on predictions, including those from LLMs. It has been applied in robotics to estimate uncertainty in learning \cite{celemin2023knowledge}, enabling systems to request clarification when confidence is low. Specific applications relevant to instruction ambiguity include ensuring probabilistically correct execution by identifying unsafe commands \cite{wang25probabilistically}, aligning LLM planner uncertainty with task requirements across various ambiguity modes (e.g., spatial, numeric) \cite{ren23robots, liang24instrospective}, and developing metrics to distinguish resolvable ambiguity from model hallucinations \cite{mullen24lap}. These works demonstrate the potential of CP to provide a principled basis for managing ambiguity by quantifying the model's certainty about its interpretation.

Although existing research addresses ambiguity from NLP foundations to general robotics, there remains a gap in systematically detecting the multiple facets of ambiguity inherent in natural language commands specifically within the high-stakes surgical context. 

\section{Methodology}

\begin{figure}[t]
\centering
\includegraphics[width=\linewidth]{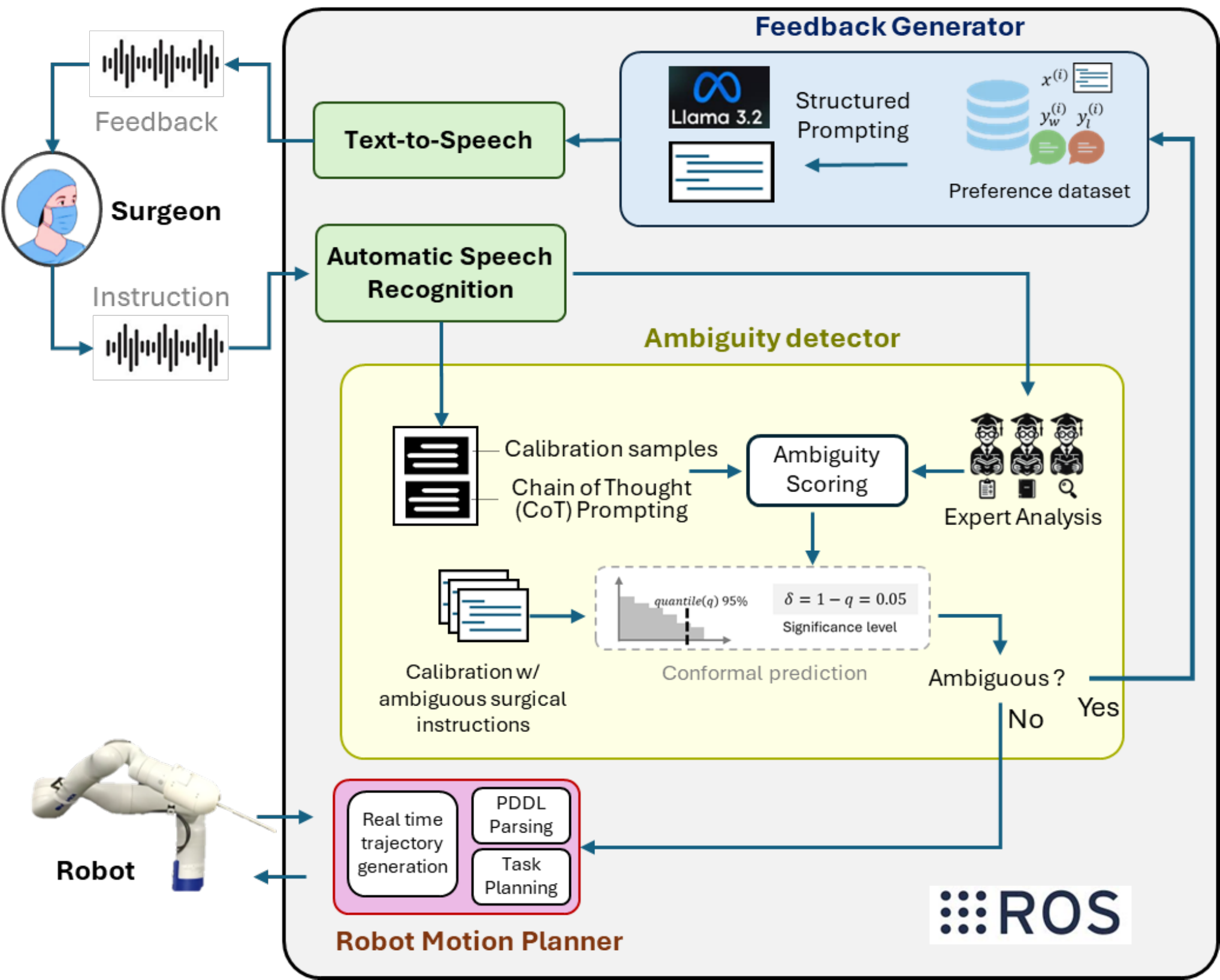}
\caption{Proposed framework for ambiguity detection}
\label{fig:2}
\end{figure}

Our framework for detecting ambiguity in natural language instructions for surgical robots integrates an ensemble of Large Language Model (LLM)-based evaluators with a conformal prediction (CP) mechanism. Each evaluator is specialized to assess different facets of potential ambiguity. The CP framework then provides a statistically rigorous classification of the instruction's clarity based on the collective assessment of the evaluator ensemble. Figure~\ref{fig:2} illustrates the overall architecture.

\subsection{Input Processing}

The system accepts natural language requests, typically originating as spoken commands in a surgical setting. As described in our previous work \cite{davila24voice}, these commands are transcribed into text using a dedicated speech-to-text module integrated via ROS, making them suitable for processing by the subsequent LLM-based pipeline.

\subsection{LLM-Based Evaluator Ensemble}
A central component is an ensemble of five distinct LLM-based evaluators designed to provide a multi-faceted assessment of instruction ambiguity. This ensemble approach leverages diverse perspectives for a more robust evaluation compared to a single, general assessment. Each evaluator utilizes a base LLM, prompted with specific instructions and illustrative examples, to generate an ambiguity score on a continuous scale from 0 (indicating complete clarity) to 10 (indicating high ambiguity).

\subsubsection{Chain-of-Thought (CoT) Evaluator}
\begin{figure}[tb]
  \centering
  \includegraphics[width=0.6\linewidth]{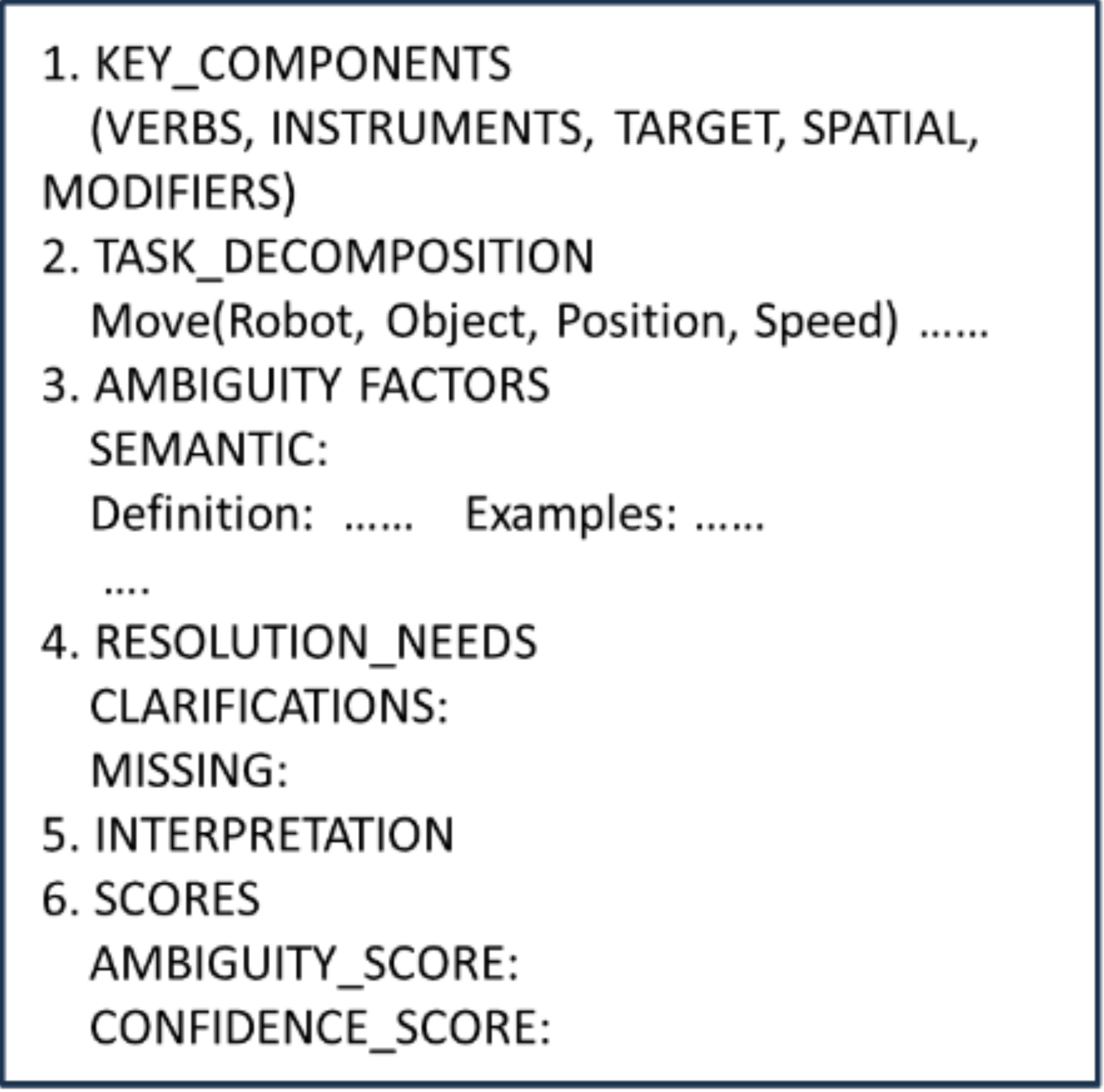}
  \caption{Structure of CoT prompt}
  \label{fig:3}
  \vspace*{-3mm}
\end{figure}

One evaluator utilizes a Chain-of-Thought (CoT) prompting strategy \cite{wei22chain}. As depicted in Figure~\ref{fig:3}, this involves guiding the LLM through a step-by-step reasoning process. The prompt is structured to first encourage the LLM to identify the key components within the instruction, such as the main verb, target objects, and any contextual elements. Following this identification, the LLM is prompted to decompose the task into a predefined list of potential robot actions relevant to surgical procedures. Subsequently, the definition and examples of multiple ambiguity factors are provided to the LLM to critically assess the potential for ambiguity based on the identified components and their relationships. These ambiguity factors include semantic, syntactic, pragmatic, and contextual types. We then request the LLM to identify potential clarifications that might be needed and any missing information required for unambiguous execution. Based on this comprehensive analysis, the LLM is asked to provide possible interpretations of the request and finally output an ambiguity score on a scale from 0 to 10, where a higher score indicates greater ambiguity. The final output of this evaluator is this ambiguity score reflecting the LLM's confidence in the instruction's clarity based on its step-by-step reasoning.

\subsubsection{Specialized Ambiguity Evaluators}
Four additional evaluators are prompt-tuned to focus on specific ambiguity types prevalent in surgical instructions:

\begin{itemize}
\item \textbf{Linguistic Evaluator:} Targets ambiguities originating from word meaning (lexical ambiguity, such as the interpretation of 'retract'), sentence structure (syntactic ambiguity, involving potentially unclear relations between phrases), or unresolved references (referential ambiguity, concerning pronouns like 'it').
\item \textbf{Contextual Evaluator:} Assesses whether the instruction is adequately grounded in the implicit or explicit surgical context. It identifies instructions that depend on unclear assumptions or lack necessary situational details.
\item \textbf{Procedural Evaluator:} Focuses on the clarity and completeness of the requested action sequence. This includes identifying underspecified parameters, exemplified by terms like 'slightly', unclear ordering of operations, or omissions of necessary steps.
\item \textbf{Critical Safety Evaluator:} Specifically screens for ambiguities with the potential to induce safety risks upon misinterpretation. Such ambiguities could involve potential tissue damage, inappropriate instrument usage, or deviations from established surgical safety protocols.
\end{itemize}

\subsection{Calibration Dataset}

The Conformal Prediction framework necessitates a calibration dataset. We constructed such a dataset comprising 40 natural language instructions representative of common surgical tasks, including endoscope navigation, instrument manipulation, and tissue handling \cite{fozilov23endoscope}, \cite{yamada23task}, \cite{liu24latent, yamada24multimodal}. Each of these 40 requests was manually labeled as either "ambiguous" or "non-ambiguous". The labeling criteria considered factors such as the clarity of the action, the specificity of the target objects, the sufficiency of the context, and the potential for misinterpretation in a surgical setting. This labeled dataset serves as the basis for calibrating our conformal predictor.

\begin{figure}[tb]
  \centering
  \includegraphics[width=0.7\linewidth]{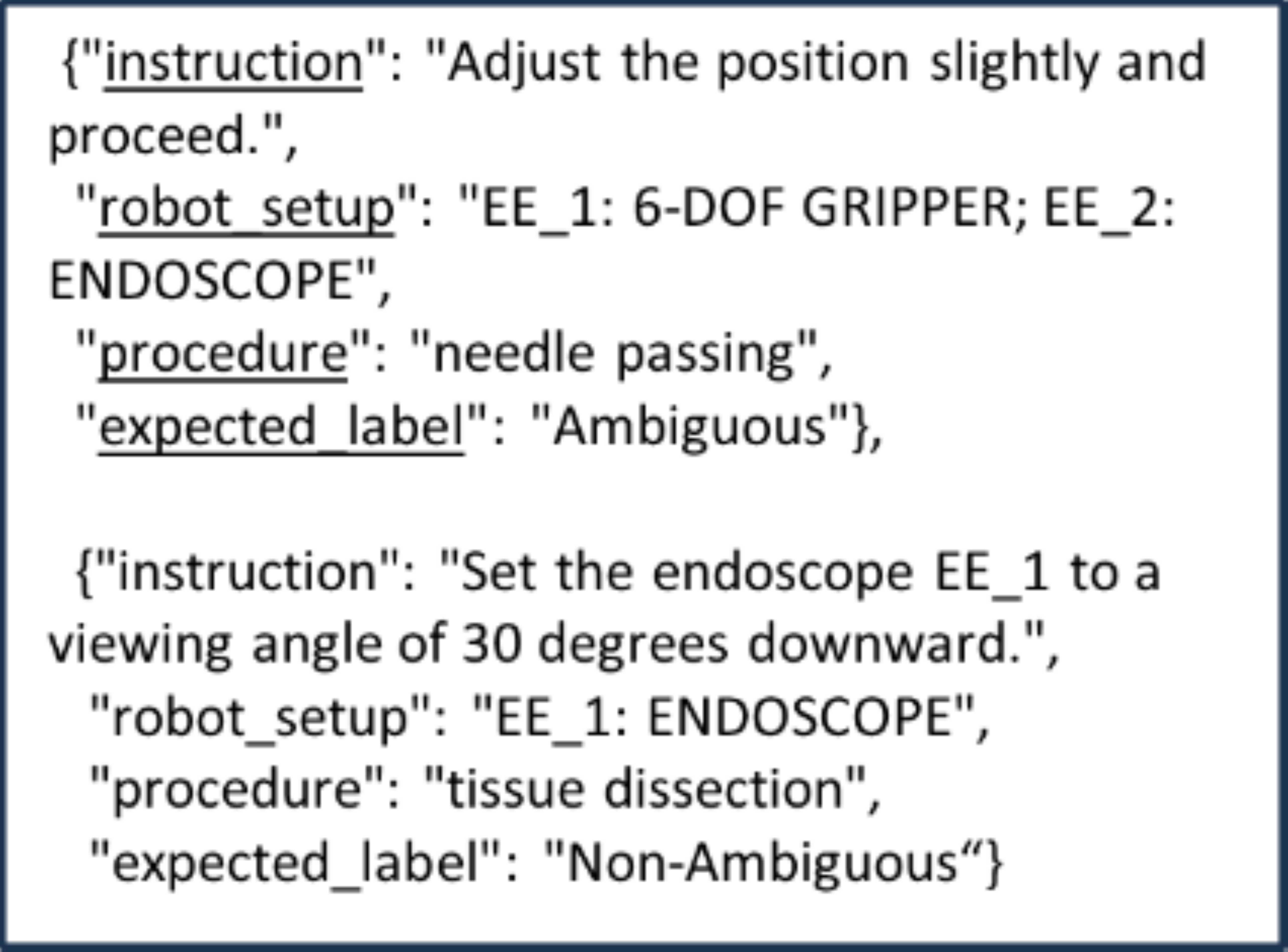}
  \caption{Examples of instructions from the calibration dataset}
  \label{fig:4}
  \vspace*{-3mm}
\end{figure}

\subsection{Conformal Prediction Framework}
We employ conformal prediction \cite{angelopoulos22gentle} to translate the ambiguity scores from the LLM ensemble into classifications with statistical validity guarantees.

\subsubsection{Ensemble Score Aggregation}
For an input instruction $i$, let $s_{i,k}$ denote the ambiguity score from the $k$-th evaluator ($k=1, \dots, 5$). We compute the mean score $\mu_i = \frac{1}{5}\sum_{k=1}^{5} s_{i,k}$ and the sample variance $\sigma_i^2 = \frac{1}{4}\sum_{k=1}^{5} (s_{i,k} - \mu_i)^2$.

\subsubsection{Nonconformity Score Calculation}
A nonconformity measure quantifies the atypicality of a new instruction $i$ relative to the calibration examples associated with a candidate class $\delta \in {\text{Ambiguous}, \text{Non-ambiguous}}$. The nonconformity score $NC_{\delta_i}$ for instruction $i$, under the hypothesis that its true class is $\delta$, is defined as:

\begin{equation} \label{eq:nc_score}
NC_{\delta_i}(\mu_i, \sigma_i^2) = |\mu_{i} - \mu_{\text{cal},\delta}| + \beta \cdot \sigma_{i}^2
\end{equation}
where:
\begin{itemize}
\item $\mu_{\text{cal},\delta}$ is the pre-computed mean of the average ambiguity scores ($\mu_j$) for all calibration examples $j$ belonging to class $\delta$.
\item $\beta \geq 0$ is a hyperparameter balancing the contributions of the mean score deviation and the score variance. A larger $\beta$ assigns higher nonconformity to instructions where the evaluators exhibited significant disagreement, reflected by a large $\sigma_i^2$.
\end{itemize}
This score is computed twice for each new instruction $i$, yielding $NC_{\text{Amb}_i}$ and $NC_{\text{NonAmb}_i}$. A low $NC_{\delta_i}$ indicates that the instruction's score profile ($\mu_i, \sigma_i^2$) is characteristic of calibration examples from class $\delta$.

\subsubsection{P-value Calculation and Classification Decision}
Utilizing the distribution of nonconformity scores from the calibration set, we compute a p-value for each hypothesis class $\delta$ for the new instruction $i$. The p-value $p_{\delta_i}$ represents the proportion of calibration examples of class $\delta$ that are at least as non-conforming as instruction $i$:

\begin{equation} \label{eq:p_value}
p_{\delta_i} = \frac{|{j \in \text{Cal}_\delta : NC_{\delta_j} \geq NC_{\delta_i}}| + 1}{|\text{Cal}_\delta| + 1}
\end{equation}

Here, $\text{Cal}_\delta$ denotes the subset of calibration examples labeled $\delta$, and $NC_{\delta_j}$ is the nonconformity score for calibration example $j$ under the hypothesis $\delta$. The addition of 1 in the numerator and denominator provides a standard adjustment for finite calibration sets.

The instruction is classified based on these p-values and a pre-specified significance level $\alpha$.

\begin{equation} \label{eq:decision}
\delta_{\text{out}} =
\begin{cases}
\text{Ambiguous} & \text{if } p_{\text{Amb}_i} > \alpha \text{ and } p_{\text{NonAmb}_i} \leq \alpha \\
\text{Non-ambiguous} & \text{if } p_{\text{Amb}_i} \leq \alpha \text{ and } p_{\text{NonAmb}_i} > \alpha \\
\text{Uncertain} & \text{if } p_{\text{Amb}_i} \leq \alpha \text{ and } p_{\text{NonAmb}_i} \leq \alpha \\
\text{Uncertain} & \text{if } p_{\text{Amb}_i} > \alpha \text{ and } p_{\text{NonAmb}_i} > \alpha
\end{cases}
\end{equation}
The primary classification outcomes are \textit{Ambiguous} and \textit{Non-ambiguous}. 

In all other cases, where there is not a clear distinction based on the p-values, we classify the instruction as \textit{Uncertain}, indicating that the system does not have sufficient statistical confidence to make a definitive classification at the chosen significance level. Both \textit{Ambiguous} and \textit{Uncertain} instructions can trigger a request for clarification from the human operator in a real-world surgical setting.

\subsection{Feedback Generation}

To provide users with actionable insights into the detected ambiguity, we implement a feedback generation module. This module utilizes a separate LLM instance, prompted to analyze the five evaluator scores ($s_{i,1}, \dots, s_{i,5}$), identify the principal ambiguity factor from the CoT evaluator, and generate a one- or two-sentence message highlighting this primary issue and suggesting how the user might clarify their instruction, for instance: "Specify the location of the tissue to be cut". This facilitates an iterative clarification process between the user and the system.

\section{Experimental Validation}

This section details the empirical evaluation of the proposed ambiguity detection framework using a curated dataset and standard classification metrics.

\subsection{Experimental Setup}
The evaluation utilized a dataset of 40 natural language instructions relevant to surgical robotics. This dataset was manually labeled and comprised 20 non-ambiguous instructions and 20 ambiguous instructions. The ambiguous set was specifically constructed to include 5 examples for each of four predominant ambiguity types: linguistic, contextual, procedural, and critical safety ambiguity. This structure allows for assessing the framework's ability to detect ambiguity overall and its sensitivity to different ambiguity manifestations.

We evaluated our framework's performance using two state-of-the-art Large Language Models (LLMs): Llama 3.2 11B and Gemma 3 12B. These models were accessed via the Transformers library from Hugging Face. All experiments were implemented in PyTorch 2.0 and conducted on a high-performance workstation featuring an AMD Ryzen Threadripper PRO 3995WX (2.7 GHz), 256 GB RAM, and an NVIDIA GeForce RTX A6000 GPU. For conformal prediction classification, the significance level ($\alpha$) was set to 0.1. The hyperparameter $\beta$, which weights the score variance in the nonconformity score calculation (Equation \ref{eq:nc_score}), was empirically determined to be 0.5 through preliminary tuning. Performance was assessed using standard classification metrics: accuracy, precision, recall, and F1-score.

\subsection{Quantitative Results}

\begin{figure}[t]
\centering
\includegraphics[width=0.9\linewidth]{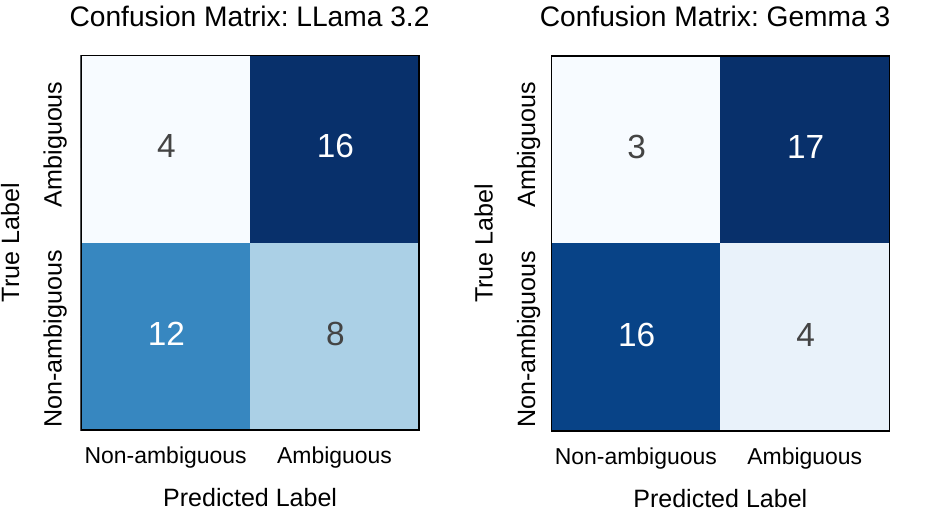}
\caption{Confusion matrices for ambiguity detection using the conformal prediction framework ($\alpha=0.1$). Left: Llama 3.2 11B (70\% accuracy). Right: Gemma 3 12B (82.5\% accuracy). }
\label{fig:5}
\end{figure}

The global performance in distinguishing between any ambiguous instruction and a non-ambiguous one is summarized by the confusion matrices in Figure~\ref{fig:5}.

\begin{figure}[t]
\centering
\includegraphics[width=0.8\linewidth]{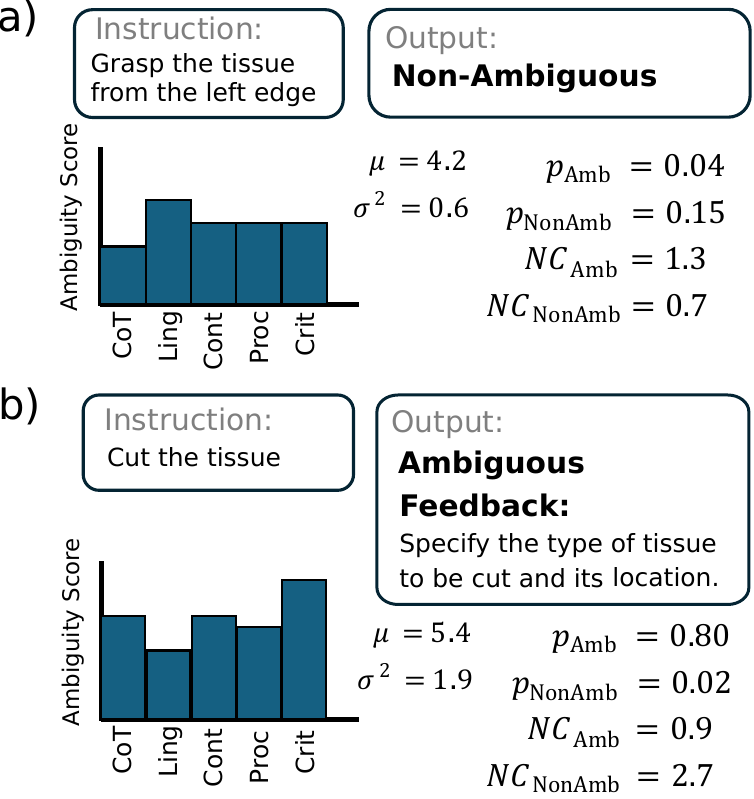}
\caption{Representative examples of the ambiguity detection process. (a) An instruction correctly classified as non-ambiguous. (b) An instruction correctly classified as ambiguous, with feedback generated based on the evaluators' outputs. The plots show individual evaluator scores and the resulting nonconformity scores (NCS).}
\label{fig:6}
  \vspace*{-3mm}
\end{figure}

Using Llama 3.2 11B, the framework achieved an overall accuracy of 70\%. It correctly identified 16 out of 20 ambiguous instructions (Recall=0.80 for ambiguous class) and 12 out of 20 non-ambiguous instructions (Recall=0.60 for non-ambiguous class). Using Gemma 3 12B, the framework demonstrated superior performance, achieving an overall accuracy of 82.5\%. This model correctly identified 17 out of 20 ambiguous instructions (Recall=0.85 for ambiguous class) and 16 out of 20 non-ambiguous instructions (Recall=0.80 for non-ambiguous class).

To understand how effectively the framework identifies specific types of ambiguity when they are present, we analyzed performance metrics focused on each category within the ambiguous set. Tables~\ref{tab:1} and \ref{tab:2} present these results. For each row, the metrics reflect the binary classification performance considering five pairs of samples, five ambiguous samples as true positives, and their corresponding five nonambiguous samples as true negatives. The "Total" row measures the performance across all 40 samples. 

With Llama 3.2 11B (Table~\ref{tab:1}), the framework correctly identified 80\% (Recall = 0.80) of the procedural ambiguity examples. Its ability to identify other types was lower, correctly identifying 60\% of linguistic and critical examples, and only 40\% of contextual examples, although precision was high for contextual cases it did flag. With Gemma 3 12B (Table~\ref{tab:2}), the framework achieved perfect recall (1.00) for linguistic ambiguities, identifying all 5 examples correctly. It also demonstrated a strong recall for contextual (0.80) and procedural (0.80) ambiguities. Performance on critical ambiguities was lower (0.60 recall), although precision was perfect, meaning it made no false positive errors when classifying critical ambiguities within that subset. The overall F1 score across all 40 samples was 0.83.

\begin{table}[tb]
\caption{Performance Metrics: Llama-3.2-11B-Vision-Instruct}
\label{tab:1}
\begin{center}
\begin{tabular}{|c||c|c|c|c|}
\hline
Ambiguity Type & F1-Score & Accuracy & Precision & Recall \\
\hline
Linguistic & 0.60 & 0.60 & 0.60 & 0.60 \\
Contextual & 0.57 & 0.70 & 1.00 & 0.40 \\
Procedural & 0.80 & 0.80 & 0.80 & 0.80 \\
Critical & 0.67 & 0.70 & 0.75 & 0.60 \\
Total & 0.73 & 0.70 & 0.67 & 0.80 \\
\hline
\end{tabular}
\end{center}
\end{table}

\begin{table}[tb]
\caption{Performance Metrics: Gemma-3-12b-it}
\label{tab:2}
\begin{center}
\begin{tabular}{|c||c|c|c|c|}
\hline
Ambiguity Type & F1-Score & Accuracy & Precision & Recall \\
\hline
Linguistic & 1.00 & 1.00 & 1.00 & 1.00 \\
Contextual & 0.73 & 0.70 & 0.67 & 0.80 \\
Procedural & 0.80 & 0.80 & 0.80 & 0.80 \\
Critical & 0.75 & 0.80 & 1.00 & 0.60 \\
Total & 0.83 & 0.83 & 0.81 & 0.85 \\
\hline
\end{tabular}
\end{center}
\end{table}

\subsection{Qualitative Analysis}

Figure~\ref{fig:6} provides illustrative examples of the framework's classification process and feedback generation.

Example (a) shows an instruction correctly identified as \textit{Non-ambiguous}, characterized by low scores from the evaluators and resulting p-values aligning with the non-ambiguous class hypothesis. Example (b) illustrates a case correctly identified as \textit{Ambiguous}. Higher scores from specific evaluators, such as the \textit{Contextual} one, contribute to p-values supporting the ambiguous classification. The feedback module then leverages the high-scoring evaluator type to generate a relevant clarification prompt for the user.

\subsection{Discussion}
The experimental evaluation confirms the viability of the proposed LLM ensemble and conformal prediction framework for detecting ambiguity in surgical instructions. The results highlight the significant influence of the underlying LLM, with Gemma 3 12B achieving considerably higher overall accuracy and more consistent detection across different ambiguity types compared to Llama 3.2 11B in our setup. This underscores the importance of model selection for tasks requiring nuanced language understanding in specialized domains.

The analysis broken down by ambiguity type reveals varying levels of sensitivity. Both models, especially Gemma 3, showed proficiency in identifying linguistic and procedural ambiguities. This suggests the framework effectively captures issues related to word meaning, sentence structure, references, and action sequence clarity. Detecting contextual and critical safety ambiguities proved more challenging, as reflected in the recall scores for these categories, particularly for Llama 3.2. This difficulty likely stems from the inherent reliance on implicit world knowledge, understanding of the ongoing surgical state, and safety constraints, which are harder for LLMs to grasp solely from the instruction text without richer context integration. 

A primary limitation of this study is the small size of the evaluation dataset, particularly the limited number of examples for each specific ambiguity type. While indicative, the results regarding type-specific performance require validation on a larger, more diverse dataset to draw robust conclusions. Furthermore, the dataset construction and labeling inherently involve subjectivity, and the framework's performance depends on the calibration set reflecting the distribution of ambiguities encountered in practice. The choice of hyperparameters ($\alpha, \beta$) and the specific nonconformity score formulation might also influence results and could be subject to further optimization.

\section{Conclusion}

In this paper, we presented a novel approach for detecting ambiguity in natural language instructions for surgical robot assistants, employing an ensemble of specialized LLM-based evaluators specializing in different ambiguity types within a conformal prediction framework. Our experimental validation using Llama 3.2 11B and Gemma 3 12B demonstrated the effectiveness of this methodology, with Gemma 3 achieving an accuracy of 82.5\% in distinguishing ambiguous from non-ambiguous instructions. Furthermore, we introduced a feedback generation mechanism to guide users in refining unclear requests. These results highlight the potential of our framework to enhance the safety and reliability of human-robot interaction in critical surgical environments. Future work will focus on expanding the calibration dataset, further refining the LLM evaluators and prompting strategies, and exploring the integration of this ambiguity detection system into a real-time surgical robotic platform.

\addtolength{\textheight}{-11cm}   







\bibliographystyle{IEEEtran}
\bibliography{biblio}

\end{document}